\RequirePackage{amsmath}
\documentclass[twocolumn]{llncs} 

\usepackage[a4paper, portrait, margin=1.2in]{geometry}
\usepackage[ruled,linesnumbered]{algorithm2e}
\usepackage{verbatim}
\usepackage{siunitx}
\SetKwRepeat{Do}{do}{while}%
\usepackage{
graphicx,
float,
multirow,
makecell,
tabularx,
booktabs,
array,
amsmath,
natbib,
hyperref
}
\usepackage[font=itshape]{quoting}
\usepackage[section]{placeins}
\usepackage{float} 
\usepackage{booktabs} 
\usepackage{url}

\emergencystretch=\maxdimen
\setcitestyle{authoryear,open={(},close={)}}

\makeatletter
\renewenvironment{abstract}{ 
    \begin{changemargin}{0cm}{0cm} 
    \setlength{\parindent}{0pt} 
    {\bfseries \setlength{\parindent}{0pt} \abstractname\vspace{\z@}}
}{\end{changemargin}}
\makeatother

\title{How to Build Robust FAQ Chatbot with Controllable Question Generator?}

\author{\large
Yan Pan\textsuperscript{a*,b*} \and
Mingyang Ma\textsuperscript{b} \and
Bernhard Pflugfelder\textsuperscript{b} \and
Georg Groh\textsuperscript{a}
}
\institute{\normalsize
\textsuperscript{a}\ Technical University of Munich, Department of Informatics,\\ Research Group Social Computing, Boltzmannstr. 3, 85748 Garching, Germany\\
\textsuperscript{b}\ BMW Group, Petuelring 130, 80788 Munich, Germany\\
\textsuperscript{*}\ corresponding author\\
}

\begin{document}
\pagestyle{plain}

\maketitle

\begin{abstract} 
Many unanswerable adversarial questions fool the question-answer (QA) system with some plausible answers. Building a robust, frequently asked questions (FAQ) chatbot needs a large amount of diverse adversarial examples. Recent question generation methods are ineffective at generating many high-quality and diverse adversarial question-answer pairs from unstructured text. We propose the diversity controllable semantically valid adversarial attacker (DCSA), a high-quality, diverse, controllable method to generate standard and adversarial samples with a semantic graph. The fluent and semantically generated QA pairs fool our passage retrieval model successfully. After that, we conduct a study on the robustness and generalization of the QA model with generated QA pairs among different domains. We find that the generated data set improves the generalizability of the QA model to the new target domain and the robustness of the QA model to detect unanswerable adversarial questions.

\keywords{
Question Generation \and Robustness \and Generalization \and Pre-trained model
}
\end{abstract}

\section{Introduction}
One goal of artificially intelligent systems is to enable the machine to answer questions from human beings. In real applications such as Azure QnAMaker \citep{agrawal2020qnamaker}, a knowledge base is built over the QA pairs given by developers and users. The AI chatbot's ability is limited by the QA pairs pre-collected in the training process, so many developed AI applications require a substantial number of high-quality QA pairs to improve the system's performance.

Manually generated QA pairs are standard datasets used to train the QA system, such as the famous benchmark SQuAD v2.0 \citep{rajpurkar2018know}. However, writing and collecting a large amount of QA pairs is time-consuming and expensive. Recently researchers used deep neural networks to automatically generate questions from knowledge base \citep{wang2020pathqg,gao2018difficulty} and knowledge graph \citep{chen2020toward, elsahar2018zero}. These generated questions are exclusively answerable questions from the same distribution in the training dataset. Since the ultimate goal is to build a reliable and robust FAQ chatbot, exploring diverse adversarial QA pairs among other domains is essential.

In the last years, the pre-trained models, e.g., BERT \citep{devlin2018bert} and GPT2 \citep{radford2019language} have achieved great success in the question generation task, which generate high diversity and quality samples \citep{klein2019learning}. Nevertheless, without control and semantic targeting, the pre-trained model may generate human-like but irrelevant sentences \citep{wu2020controllable}. ACS-QG \citep{liu2020asking} model proved that people would consider various factors to ask a question, such as an answer, style, and clues. Therefore besides the question qualities, the generation model is controlled by semantic diversity and difficulty labels \citep{gao2018difficulty}. 

The FAQ chatbot always consists of passage retrieval and reading comprehension modules to answer questions over knowledge sources. Efficient passage retrieval in question answering is usually implemented using keyword matching, like TF-IDF or BM25 \citep{robertson2009probabilistic}.  In this paper, we choose the TF-IDF-based defacto retrieval model \citep{chen2017reading} to simulate retriever processing in the industry to handle a large number of documents \citep{agrawal2020qnamaker}. Moreover, the reading comprehension model has improved rapidly and outperforms the human's performance in some benchmarks. Unlike the most current work focused on the model, we are interested in improving the robustness and generalization with data augmentation across other domains.

To conclude, we use a controllable question generation model with a semantic graph to generate diverse adversarial examples. After syntactic and semantic filtering, retraining and transfer learning are implemented among different challenge sets. Our findings and contributions are three-fold:

\begin{itemize}
\item[$\bullet$] A semantic and syntactic controllable  model to generate diverse fluent question answering pairs with a semantic graph;
\item[$\bullet$] An analysis of adversarial questions over retriever and reading comprehension model;
\item[$\bullet$] Improved robustness and generalization with generated adversarial examples among different domains.
\end{itemize}

\section{Related Work} \label{sec:theory}

Question generation has the potential to improve the training of QA systems \citep{du2018harvesting,tang2017question}, and help chatbots start a conversation with human users \citep{mostafazadeh2016generating}. However, the well-designed templates in rule-based lack diversity, and they do not generalize to a new domain \citep{ali2010automatic}. The recent neural-based models, like the encoder-decoder structure with attention mechanism, perform better and attract much attention in the research field \citep{du2018harvesting}. The answer information \citep{kim2019improving} and lexical features are incorporated into the generation model without hand-crafted templates. However, the irrelevant and uninformative generated questions are worthless in the future QA system  \citep{gao2018difficulty}.

Regarding the robustness of the QA system, adversarial attacks are designed to fool the deep learning models \citep{zhang2020adversarial}, including white-box \citep{liang2017deep} and black-box attacks \citep{li2020bert}. For the reading comprehension task, models can not answer the questions correctly with the adversarially inserted sentences \citep{jia2017adversarial}. The unanswerable questions with plausible answers are also written adversarially in the SQuAD v2.0 \citep{rajpurkar2018know}. \citet{sen2020models} analyzed variant datasets and suggestions in the future dataset building. \citet{talmor2019multiqa} and \citet{shinoda2021improving} argued that multiple source datasets improve the QA model's performance and robust generalization. 

The reading comprehension model combined with a search component can tackle question answering \citep{chen2017reading}. A retrieval system with TF-IDF/BM25 is widely implemented in real-world applications. In advance of pre-trained model, Roberta \citep{liu2019roberta} and Albert \citep{lan2019albert} work successful as reader component.

\section{Methodology} \label{sec:method}
\begin{figure*}
\centering
\includegraphics[scale=0.55]{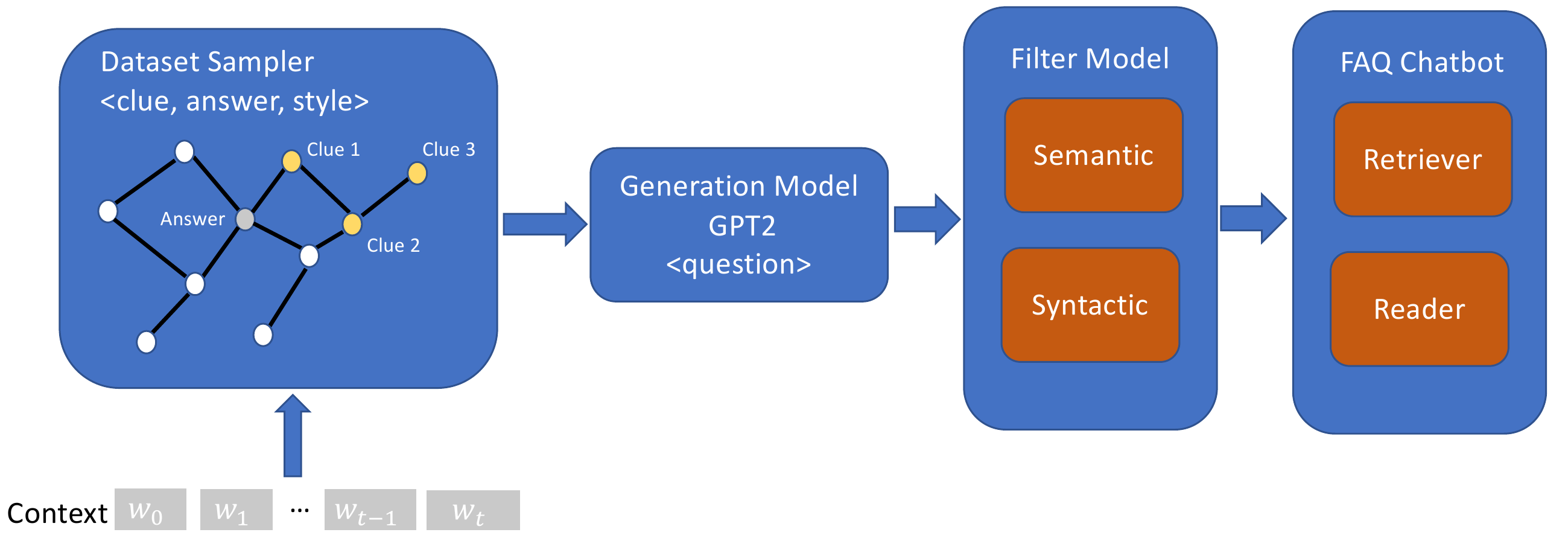}
\caption{\textbf{Overall system architecture} includes dataset sampler, generation model, question filter model, and FAQ chatbot. Input is the unstructured context consisting of $t$ words $w_1,w_2,...,w_t$}
\label{fig:overall system architecture}
\end{figure*}

Figure \ref {fig:overall system architecture} shows our overall system architecture for generating diverse, high-quality questions with a semantic graph. Our system consists of four components: \romannumeral1) datasets sampler, which recognizes the facts and relationship as symbolic presentations with a semantic graph and samples the answer, style, clues for controllable diverse question generation; \romannumeral2) high-quality question generation model, which is fine-tuned on the constructed data to generate amounts of human-like questions; \romannumeral3) question filters, which filter out the wished samples based on semantic and syntactic features; \romannumeral4) FAQ chatbot, which evaluates the quality of adversarial examples in the QA task.

\subsection{\textbf{Parsing Answer, Style and Clues for Question Generation}}

The first main task is to acquire all candidate facts, including the answer, style, and clues based on a given passage. The sampler mines various candidates from the passages with SceneGraphParser \citep{wu2019unified}. We assume that question-worthy facts, clues, and answers over the knowledge context rely on dependency parsing. The SceneGraphParser parses sentences into scene graphics, where nodes and edges are symbolic representations. We get all candidate nodes by parsing and named entity recognition over the passage.

Inspired by PathQG \citep{wang2020pathqg} and ACS-QG \citep{liu2020asking}, we select multiple clues and answer from the semantic graph and create the triples with all possible styles. It makes sense because multiple QA pairs can be extracted from different aspects. Furthermore, we evaluate the relationship between the clues over the graph. The nearby clue nodes simulate the various facts when a human asks questions and guarantee semantic consistency.  

\subsection{\textbf{GPT2-Based Question Generation}}

In comparison to Seq2Seq models, the power of the pre-trained language model GPT2 \citep{radford2019language} has been proved in the language generation task. The large pre-training dataset of GPT2, which is accessible through Huggingface \footnote{https://huggingface.co/}, makes it easier to generate diverse samples. We use the ACS-aware question generation model, where the semantic control over GPT2 is ensured with the concatenated passage, answer, clues and style. Moreover, we train the GPT2 model based on both standard answerable questions and unanswerable adversarial questions without an answer. The number of inserted clues determines the question information.

\subsection{Semantic and Syntactic Filter Model}
Based on the ACS-aware model, we could generate human-like fluent diversity adversarially questions. However, the questions can be meaningless, especially for the fine-tuned QA system. One task of the filter model is to enhance the semantic validity between the questions and answers. So we leverage an existing QA model to control the semantic quality.

Besides semantic meaning, we further utilize a TF-IDF-based filter to control the syntactic quality. First, the TF-IDF is the core algorithm applied in the retriever with a large passage knowledge base. Many paraphrasing questions are lost in retriever processing, so we call the answerable questions with low TF-IDF syntactic scores adversarially attacks for the retriever system. Second, reader models tend to answer questions with high word overlap with the context \citep{sen2020models}, so the selected unanswerable questions with over TF-IDF threshold scores are valuable adversarial samples for the reader component.

\subsection{TF-IDF-Based Retriever and Roberta-Based Reader}
Our final target is to improve the FAQ chatbot's performance and evaluate the generated questions on the downstream tasks. Therefore we apply the generated datasets to both retriever and reader.

We achieve the black-box attack by sending the selected questions to the target FAQ system. In order to validate the improvement of robustness and generalization, a Roberta-based model \citep{liu2019roberta} is trained on the different challenge sets with adversarial examples included. 

\section{Experiments} \label{sec:results}
This section introduces the data and the basic setup we used. The QA generation models create synthetic datasets. Moreover, the robustness, generalization, and accuracy of the FAQ system are evaluated and compared among different challenge sets. 

\subsection{Question Answering Datasets}
We use the popular span based benchmark dataset SQuAD v1.0 \citep{rajpurkar2016squad}, SQuAD v2.0 \citep{rajpurkar2018know}, QuAC \citep{choi2018quac}, and NQ \citep{kwiatkowski2019natural}  for question answering task following \citep{sen2020models}. The SQuAD datasets are used for our proposed model's training. Question/answer/passage triples are provided by each dataset, including answerable and unanswerable questions. Below we briefly describe the four datasets, and Table \ref{tab:train-dataset} shows the detailed statistic information of the datasets.

\begin{table}
    \centering
    \setlength{\abovecaptionskip}{5pt}
    \setlength{\belowcaptionskip}{-30pt}
    \setlength{\tabcolsep}{4.0mm}{
    \begin{tabular}{c|c|c}
        \toprule
        Dataset & Train & Dev \\
        \midrule
        SQuAD v1.0  & 87,599 & 10,570  \\ 
        SQuAD v2.0  & 130,319 & 11,873 \\ 
        QuAC & 83,568 & 7,354  \\ 
        NQ & 110,857 & 3,368  \\ 
        \bottomrule
    \end{tabular}
    \caption{Train and dev set sizes of the QA datasets, SQuAD v1.0, SQuAD v2.0, QuAC, and NQ }
    \label{tab:train-dataset}}
\end{table}
\subsubsection{SQuAD} Stanford Question Answering Data Set v1.0 is a high-quality English reading comprehension data set. The released SQuAD v2.0 in 2018 has a higher robustness requirement with unanswerable adversarial questions. The relevant questions and plausible answers make it more challenging to distinguish between adversarial questions. \citet{yatskar2018qualitative} argued that the diversity of unanswerable questions is higher than other datasets. So the SQuAD 2.0 dataset contains rich adversarial language features, which is crucial in our DCSA generation model.

\subsubsection{QuAC} QuAC is a large dialog-style dataset that involves two annotators. The answer comes from the spans in the evidence text, constructed from Wikipedia articles by source crowd workers. In comparison to the SQuAD dataset, many similar principles are shared within QuAC, but it also brings more challenges, including unanswerable questions and dialog context \citep{choi2018quac}.

\subsubsection{Natural Questions (NQ)} NQ dataset consists of 300K examples issued to Google search engine, representing real information from people. We make use of the high-quality annotations of long and short answers from each sample.

\subsection{Experiment Settings and Automatic Evaluation}
First, we conduct the QG experiment on the SQuAD dataset, where outputs are some questions. GPT2-ACS \citep{liu2020asking} is used as our baseline. During the decoding process, the top-p sampling suggested by \citet{liu2020asking} is implemented with beam search. We fine-tune the Roberta-base QA model from HuggingFace for up to 2 epochs with batch size 32. The training parameters are following the experiment setting in the Haystack \footnote{https://github.com/deepset-ai/haystack/}. Based on the requirement and properties of our black-box adversarial attacks, we evaluate the generated questions in four aspects: syntactic diversity, semantic validity, fluency, and accuracy.

\subsubsection{Semantic Validity} \label{sec:CLC}

In the previous section, we show that the questions should be relevant to the correlated passages. We use the pre-trained multilingual universal sentence encoder \citep{chidambaram2018learning}, an embedding-based similarity metric to evaluate the semantic validity. The correlated passages have a high dot product similarity score with the questions they answer \citep{yang2019multilingual}. 

\subsubsection{Syntatic Diversity} 

Entropy-4 (Ent-4) \citep{zhang2018generating} have been widely used to measure syntactic diversity. We use average entropy and average question length ($|$U$|$) to show the diversity and information content of generated questions \citep{serban2017hierarchical}. Meanwhile, the existing word-level overlapping metrics BLEU and ROUGE-L are meaningless as we aim to diversify the generated questions with external clues.

\subsubsection{Fluency} \label{sec:CLC}
The GPT2 from Huggingface uses the Perplexity (Perp) to evaluate the fluency of generated questions. Because we aim to generate human-like queries, lower perplexity means a better QA generation.

\subsubsection{Accuracy} \label{sec:CLC}
To evaluate the QA model's performance benefit from generated questions, we use F1 and EM scores following the SQuAD and QuAC. Compared to the extract-match metric, the F1 score of word overlap is more suitable for our task due to the answer flexibility.
\subsection{Evaluating Results} \label{sec:CLC}
Because the SQuAD test set is not released, we utilize the SQuAD dataset split for a fair comparison in QG experiments, following \citet{zhou2017neural} for a fair comparison in QG experiments. The results in Table \ref{tab:Evaluation on the result} show that our model can generate fluent questions with the lower perplexity of 5.59. The generated questions from the DCSA model have higher semantic similarity with input context than the baseline model. Moreover, the average entropy and question length imply that the semantic graph improves diversity and increase the information content of generated question while choosing clue paths compared to GPT2-ACS. 

\begin{table}
    \centering
    \setlength{\abovecaptionskip}{5pt}
    \setlength{\belowcaptionskip}{-21pt}
    \setlength{\tabcolsep}{0.7mm}{
    \begin{tabular}{p{0.25\linewidth}|p{0.22\linewidth}|p{0.15\linewidth}|p{0.15\linewidth}|p{0.1\linewidth}}
        \toprule
        Generation model & Semantic & $|$U$|$  & Ent-4 & Perp  \\
        \midrule
        GPT2-ACS & 0.46 & 10.28 & 11.05 & 6.01\\
        DCSA & 0.57 & 12.26 & 11.15 & 5.59\\
        
        \bottomrule
    \end{tabular}
    \caption{Evaluation results of QG generation on the SQuAD test set \citep{zhou2017neural} }
    \label{tab:Evaluation on the result}}
\end{table}
\subsection{Human Evaluation} \label{sec:He}

Due to the surface-form limitation of commonly used evaluation metrics, we analyze the quality of the generated adversarial QA pairs with human evaluation. Three volunteers score randomly selected 100 question-passage samples generated by GPT2-ACS and our model. We also choose 50 samples from the SQuAD test set. We ask the annotators to label the example from 0-2 following three questions: (a) if the question is relevant to the context; (b) if the question is grammatically correct and fluent; (c) if the question is informative. 
\begin{table}
    \centering
    \setlength{\abovecaptionskip}{5pt}
    \setlength{\belowcaptionskip}{-2pt}
    \setlength{\tabcolsep}{3mm}{
    \begin{tabular}{p{0.35\linewidth}|p{0.10\linewidth}|p{0.10\linewidth}|p{0.10\linewidth}}
        \toprule
        Dataset & Rel & Flu & Inf\\
        \midrule
        GPT2-ACS & 1.25 & 1.71  & 1.45 \\ 
        DCSA & 1.45  & 1.80 & 1.55\\ 
        \midrule
        SQuAD ori & 1.70  & 1.90 & 1.79 \\ 
        \bottomrule
    \end{tabular}
    \caption{The relevant, fluent, and informative scores of questions from human-evaluation}
    \label{tab:human evaluation}}
\end{table}

The results in Table \ref{tab:human evaluation} show that our generated questions appear more relevant to the context's topic with a score of 1.45 than the GPT2-ACS model. According to the fluent score, questions from both models are human-like fluent. However, the DCSA model can generate more precise and informative questions. 

\subsection{Attacking Retriever Results} \label{sec:CLC}

We train the DCSA model with SQuAD v2.0 datasets to generate more adversarial questions for the retrieval and reader model. We select the questions that have high semantic scores with context and less word matching. As shown in Table \ref{tab:Attacking result}, the selected adversarial samples successfully fool the DrQA retriever system \citep{chen2017reading}. The syntactic-based passage retriever fails to match the correct passage and answer. Compared to DrQA, the top1 accuracy decreases $17\%$.
\begin{table}
    \centering
    \setlength{\abovecaptionskip}{5pt}
    \setlength{\belowcaptionskip}{-16pt}
    \setlength{\tabcolsep}{0.5mm}{
    \begin{tabular}{p{0.5\linewidth}|p{0.2\linewidth}|p{0.25\linewidth}}
        \toprule
         Syntatic score &Retriever& ACC Score\\
        \midrule
         SQuAD v2.0 &DrQA & 74.77 \\ 
         Generated &DrQA &57.39  \\
        \bottomrule
    \end{tabular}
    \caption{Retriever accuracy of original and generated SQuAD v2.0 dataset}
    \label{tab:Attacking result}}
\end{table}

\subsection{Generalization and Robustness} \label{sec:CLC}

One of our final goals is to improve the QA model's performance in a new domain. So we examine generalization and robustness by asking two questions: (a) if models benefit from generated QA samples while generalizing to a new domain. (b) if the adversarial training with generated unanswerable questions helps the QA model become robust.

\begin{table}
    \centering
    \setlength{\abovecaptionskip}{5pt}
    \setlength{\belowcaptionskip}{-2pt}
    \setlength{\tabcolsep}{0.5mm}{
    \begin{tabular}{p{0.30\linewidth}|p{0.098\linewidth}|p{0.098\linewidth}|p{0.098\linewidth}|p{0.098\linewidth}|p{0.098\linewidth}|p{0.098\linewidth}}
        \toprule
          & \multicolumn{2}{c|}{SQuAD} & \multicolumn{2}{c|}{QuAC} &  \multicolumn{2}{c}{NQ}\\
        \midrule
        Training & F1 & EM & F1 & EM & F1 & EM \\
        \midrule
        SQuAD & 83.18  & 80.05  & 22.42  & 12.08 & 52.64 & 46.87\\
        +SQuAD ge & 84.02 & 80.94 & 23.32 & 12.92  & 53.01 & 47.22 \\
        +QuAC ge & 83.78 & 80.49  &  25.53 & 16.37 & 51.69 & 46.48 \\
        +NQ ge & 83.58 & 80.36 & 24.10 & 14.35 & 53.12 & 46.87 \\
        \bottomrule
    \end{tabular}
    \caption{Compare F1 and EM scores on the validation set for the target dataset. Rows corresponded to the different training datasets, including the original dataset and generated dataset (ge). And columns show the evaluation results over the target dataset.  }
    \label{tab:Retraining of Question Answering}}
\end{table}

To answer the above two questions, the Roberta model is fine-tuned based on different settings, including the source dataset, generated source dataset, and generated target dataset. We evaluate F1 and EM scores against all three test sets, and the results are shown in Tables \ref{tab:Retraining of Question Answering} and \ref{tab:Adversarial Result}. Taking a closer look, we observe that the model trained with the target generated dataset achieves better performance than with the original dataset alone when generalizing to the target domain. It proves that the data augmentation on the target set helps models generalize to the new domain and learn the new knowledge.

\begin{table}
    \centering
    \setlength{\abovecaptionskip}{5pt}
    \setlength{\belowcaptionskip}{-2pt}
    \setlength{\tabcolsep}{0.4mm}{
    \begin{tabular}{p{0.58\linewidth}|p{0.15\linewidth}|p{0.1\linewidth}|r@{.}p{0.04\linewidth}}
        \toprule
         Train     & F1 & ans & \multicolumn{2}{c}{unans}\\ 
        \midrule
        SQuAD generated ans & 44.33 & 87.23 & 1&55  \\
        SQuAD generated ans+unans & 77.11 & 73.83 & 80&37 \\
        \bottomrule
    \end{tabular}
    \caption{Compare F1 scores on the validation set of the SQuAD v2.0 dataset. Rows correspond to the different fine-tuned Roberta models, and columns show the evaluation results based on SQuAD v2.0 dev datasets' chosen part, including answerable and unanswerable questions. }
    \label{tab:Adversarial Result}}
\end{table}
Second, the Roberta model becomes more robust owing to the power of generated adversarial unanswerable questions. We discover that the capability to detect unanswerable questions is generally improved from 1.55\% to 80.37 \% in Table \ref{tab:Adversarial Result}. It implies that generating unanswerable adversarial questions improved adversarial robustness.

\subsection{Accuracy} \label{sec:He}

Table \ref{tab:Retraining of Question Answering} shows that the QA model trained by the original SQuAD v2.0 set gives 83.18\% F1 and 80.05\% EM. In comparison, we can further improve the performance with additionally generated triples from the same domain. When combining the generated sets with the original dataset as training data, the implementation gives 84.02\% F1 and 80.94\% EM.

We also find that out-of-domain generated samples lead to better performance than training on source domain alone. The F1 score on SQuAD v2.0 dev set increases from 83.18\% to 83.78 \% with generated QuAC dataset and from 83.18\% to 83.58 \% with generated NQ dataset. The result is reasonable because out-of-domain generated triples bring new language knowledge and enhance the model's generalization capability.

\section{Conclusion} \label{sec:Conclusions}
This paper proposes a DCSA model to generate answerable and unanswerable questions combined with a semantic graph from raw documents. We simulate the way when humans ask questions as the nodes in the semantic graph. And the semantic and syntactic filters are utilized to sample the valuable adversarially triples according to retriever and reader models.

We present and evaluate the generated samples from semantic, syntactic, and fluency aspects. Compared with the existing question generation model, we analyze the generalization and robustness benefits among different domains. The generated out-of-domain samples can enhance the reader model's capability and are very helpful to improve the QA model's domain adaptation performance.

\section*{CRediT authorship contribution statement}
\textbf{Yan Pan:} Conceptualization, Methodology, Validation, Formal analysis, Investigation, Writing-original draft, Writing-review\&editing, Visualization, Project administration. 
\textbf{Mingyang Ma:} Conceptualization, Methodology, Validation, Writing-review, Supervision, Project administration. 
\textbf{Bernhard Pflugfelder:} Conceptualization, Methodology, Writing-review, Supervision, Project administration, Funding acquisition. 
\textbf{Georg Groh:} Conceptualization, Writing-review, Supervision, Project administration.

\section*{Acknowledgements}
Funding: This work was supported by the BMW Group. We thank Andrea for his helpful comments.

\bibliography{literature}

\newpage

\end{document}